%% file: neurips_distshift_2023.tex
\documentclass[final]{article}

\usepackage{sidecap}

\usepackage[nonatbib, final]{neurips_distshift_2023}

\usepackage{graphicx}
\usepackage{amsmath}
\usepackage{amssymb}
\usepackage{booktabs}
\usepackage{multirow}
\usepackage{bbm}
\usepackage{amsmath}
\usepackage{xcolor}
\usepackage{wrapfig}
\usepackage{caption}
\usepackage{threeparttable}

\usepackage{blindtext}

\DeclareMathOperator{\argmax}{argmax}

\DeclareMathOperator{\Dis}{\mathtt{Dis}}
\DeclareMathOperator{\dis}{\mathtt{dis}}

\DeclareMathOperator{\error}{\mathtt{Err}}

\usepackage{glossaries}
\newacronym{gde}{GDE}{Generalization Disagreement Equality}
\newacronym{ood}{OOD}{out-of-distribution}
\newacronym{id}{ID}{in-distribution}
\newacronym{kld}{KLD}{Kullback-Leibler divergence}
\newacronym{mape}{MAPE}{mean absolut percentage error}
\newacronym{msp}{MSP}{maximum softmax probability}
\newacronym{mlgt}{MaxLogit}{maximum logit}
\newacronym{hd}{HD}{Hellinger distance}
\newacronym{tvt}{TvT}{testbed vs. testbed}
\newacronym{tvc}{TvC}{testbed vs. CLIP}
\newacronym{fm}{FM}{foundation model}
\newacronym{jsd}{JSD}{Jensen–Shannon divergence}

\glsdisablehyper

\definecolor{agreement}{rgb}
{0.08419838523644754, 0.31280276816609, 0.5534025374855825}
\definecolor{accuracy}{rgb}
{0.6, 0.06274509803921569, 0.15294117647058825}
\definecolor{top1_grey}{RGB}
{119,119,119}
\definecolor{kl_green}{RGB}
{170,204,159}
\definecolor{hd_blue}{RGB}
{125,169,176}
\usepackage[pagebackref,breaklinks,colorlinks]{hyperref}
\hypersetup{colorlinks,linkcolor={top1_grey},citecolor={top1_grey},urlcolor={top1_grey}}  

\usepackage[utf8]{inputenc} 
\usepackage[T1]{fontenc}    
\usepackage{hyperref}       
\usepackage{url}            
\usepackage{booktabs}       
\usepackage{amsfonts}       
\usepackage{nicefrac}       
\usepackage{microtype}      
\usepackage{xcolor}         

\usepackage[capitalize]{cleveref}
\crefname{section}{Sec.}{Secs.}
\Crefname{section}{Section}{Sections}
\Crefname{table}{Table}{Tables}
\crefname{table}{Tab.}{Tabs.}

\title{Beyond Top-Class Agreement: Using Divergences to Forecast Performance under Distribution Shift}

\author{%
  Mona Schirmer \\
  UvA-Bosch Delta Lab \\
  University of Amsterdam\\
  \And
    Dan Zhang \\
   Bosch Center for AI \& \\
   University of Tübingen \\
   \And
   Eric Nalisnick \\
   UvA-Bosch Delta Lab \\
   University of Amsterdam \\
}

\begin{document}

\maketitle

\begin{abstract}
Knowing if a model will generalize to data `in the wild' is crucial for safe deployment. To this end, we study model disagreement notions that consider the full predictive distribution - specifically disagreement based on Hellinger distance, Jensen-Shannon and Kullback–Leibler divergence.  We find that divergence-based scores provide better test error estimates and detection rates 
on out-of-distribution data compared to their top-1 counterparts. Experiments involve standard vision and foundation models.
\end{abstract}

\section{Introduction}


\label{sec:intro}
\input{plots/divergence}

When deployed to the wild, machine learning systems often encounter conditions unlike those on which the system was trained. These deviations from the training distribution pose significant risks in safety-critical applications, such as autonomous driving, since performance can suddenly and precipitously degrade. It is therefore essential to develop techniques that can anticipate system failures caused by operating under distribution shift.  However, estimating test-time performance is difficult in even \gls{id} settings.  \Gls{ood} scenarios are especially challenging since a labeled \gls{ood} validation set is often impossible to obtain. 


To help with this problem, recent studies \cite{nakkiran2020distributional, jiang2021assessing, baek2022agreement, kirsch2022note, rosenfeld2023almost} have identified strong empirical correlations and theoretical connections between test error and model disagreement. \cite{nakkiran2020distributional,jiang2021assessing} found that model disagreement rates serve as an effective approximation of the test error and  \cite{jiang2021assessing, kirsch2022note} show that the two are equivalent under calibration.
Moreover, \cite{baek2022agreement} found that \gls{id} vs. \gls{ood} disagreement correlates linearly and the slope and bias match those of ID vs. OOD test error. 
These discoveries motivate the use of model disagreement for unlabeled OOD performance estimation.

Previous approaches to estimate a model's OOD accuracy are often based on either the model's uncertainties \cite{hendrycks2016baseline,garg2022leveraging, lu2023predicting} or 
top-1 disagreement with other models \cite{jiang2021assessing,baek2022agreement, rosenfeld2023almost}. 
In this paper, we ask a natural next question; can we combine the best of both worlds by using model disagreement notions that consider the full predictive label distribution? 
Such distributional disagreement notions have already found applications in e.g. active learning \cite{nigam1998employing, melville2005active, dvornik2019diversity} and uncertainty quantification \cite{malinin2020uncertainty}. However, its role for performance estimation has been unexplored.
We argue that the use of the predictive distribution provides a more nuanced view of model disagreement as visualized in \cref{fig:divergence}. By investigating three classic divergences, we show their value in OOD detection and error estimation.

\section{Preliminaries}
\label{sec:background}

\vspace{-5pt}

\paragraph{Problem formulation} We consider a classification task with a set of classifiers $\mathcal{F}$ where $f \in \mathcal{F}$ maps from the feature space $\mathcal{X} \subseteq \mathbb{R}^D$ to the probabilities over $K$ classes, i.e. $f: \mathcal{X} \to [0,1]^{K}$. 
The classifiers have been trained on a training set $\mathcal{S}_{ID} = \{(x_i, y_i)\}_{i=1}^{N} $ with $x_i \in \mathcal{X}$ and  $y_i \in \mathcal{Y}=[K]$ sampled from the in-distribution $\mathcal{D}_{ID}$. 
In this paper, we ask the question what is a good measure of model disagreement $\dis$ that can help to infer the 
model performance on an unseen test set $\mathcal{T}_{OOD} = \{(x_i, y_i)\}_{i=1}^{N'} $ sampled from a shifted distribution $\mathcal{D}_{OOD}$. To have a consistent translation from error to disagreement for each notion, we denote as error the respective disagreement between the model and the one-hot encoded class label, $\Tilde{y}$. The disagreement of two classifiers $f, f'$ as well as error for a data distribution $\mathcal{D}$ is then given by 
$$
\Dis_{\mathcal{D}}(f, f') = \mathbb{E}_{x,y\sim \mathcal{D}}[\dis(f(x), f'(x))], 
\quad \textrm{and} \quad 
\error_{\mathcal{D}}(f) = \mathbb{E}_{x,y\sim {\mathcal{D}}}[\dis(\Tilde{y}, f(x))].
$$

We can estimate $\Dis_{\mathcal{D}}$ and $\error_{\mathcal{D}}$ as the empirical average over samples from $\mathcal{D}$. Note that while test error requires access to labels, disagreement can be computed using only unlabeled samples.




\vspace{-5pt}
\paragraph{Previous approaches}  Past work on the link between agreement and generalization \cite{jiang2021assessing,kirsch2022note,baek2022agreement, rosenfeld2023almost} has focused on top-1 disagreement to predict test error rates. It states the disagreement rate of two classifiers on the predicted class label. Let $h: [0,1]^{K} \to [K]$ denote the function that converts the output probabilities to the predicted class label, i.e. $h(f(x)) = \argmax\limits_{k \in [K]}  f_k(x)$ where $ f_k(x)$ denotes the predicted probability for class $k$. Then, top-1 disagreement and test error is defined by 
$$
\dis^{Top1}(f(x),f'(x)) = \mathbbm{1}\{h(f(x))   \neq h(f'(x))\}
\quad \textrm{and} \quad 
\error_{\mathcal{D}}^{Top1}(f) = \mathbb{E}_{x,y\sim {\mathcal{D}}} [\mathbbm{1}\{h(f(x))   \neq y\}].
$$



\section{Divergence Disagreement}

In this work, we investigate divergence-based disagreement notions arguing that the previous top-1 notion is too coarse to differentiate between in fact different disagreement scenarios. Consider this motivating example: Three multi-class classifiers $A$, $B$, and $C$ predict class $k$ for a given sample. Model $A$ and $B$ predict class $k$ with probability $0.99$ and model $C$ with $0.2$. Under top-1 disagreement, all models would be considered in agreement despite the clear alignment difference between models A and B versus model C. 
By considering a model's predictive distribution, we can define disagreement notions that differentiate between \textit{discrepancies in uncertainties} across models. We examine disagreement notions based on three commonly used divergences: Hellinger distance, Jensen-Shannon and Kullback-Leibler divergence.

\paragraph{\gls{hd} disagreement} 
\input{plots/divergence_error}
The Hellinger distance is symmetric and satisfies the triangle inequality rendering it a true metric on the space of probability distributions. Notably, it lies in the interval $[0,1]$ making it straight forward for comparison. \gls{hd} disagreement and test error with respect to the true class label $y$ are given by 
$$
\dis^{HD}(f(x), f'(x)) =
\frac{1}{\sqrt{2}}
\sqrt{\sum_{k} \Big(\sqrt{f_k(x)} - \sqrt{f_k'(x)}\Big)^2}
$$
$$
\error_{\mathcal{D}}^{HD}(f) = 
    \mathbb{E}_{x,y\sim \mathcal{D}}
    \Bigg[
    \frac{1}{\sqrt{2}}
    \sqrt{\big(\sqrt{f_{y}(x)} - 1\big)^2
    + \sum_{k\neq y} f_k(x)}
    \Bigg]
$$

\paragraph{\gls{jsd} disagreement} \gls{jsd} is a symmetric divergence based on the \gls{kld}. It averages the \gls{kld} of their arguments from their uniform mixture and is bounded by $\log 2$. The \gls{jsd} disagreement notion is given by 
$$
\dis^{JSD}(f(x), f'(x)) = 
    \frac{1}{2} \Big( \sum_{k}f_k(x) \log \frac{f_k(x)}{\overline{f_k}(x)} 
    + \sum_{k}f'_k(x) \log \frac{f'_k(x)}{\overline{f_k}(x)}\Big).
$$
\text{with} $\overline{f_k}(x) = \frac{1}{2}(f_k(x) + f'_k(x))$ denoting the uniform mixture. The \gls{jsd} error can be formulated as 
$$
\error_{\mathcal{D}}^{JSD}(f) = 
    \mathbb{E}_{x,y\sim \mathcal{D}}
    \Bigg[
    \frac{1}{2}\Big(
    \log \frac{2}{1+f_y(x)}
    +f_y(x)\log \frac{2f_y(x)}{1+f_y(x)}
    +\sum_{k\neq y}f_k(x)\log2
    \Big)
    \Bigg].
$$

\paragraph{\gls{kld} disagreement}
Unlike \gls{hd} and \gls{jsd}, the standard \gls{kld} is non-symmetric and unbounded. We consider a simple symmetrized version of the \gls{kld} previously employed in active learning \cite{dvornik2019diversity} which averages over forward and reverse \gls{kld}, 
$$
\dis^{KLD}(f(x), f'(x)) = 
    \frac{1}{2} \Big( \sum_{k}f_k(x) \log \frac{f_k(x)}{f'_k(x)} 
    + \sum_{k}f'_k(x) \log \frac{f'_k(x)}{f_k(x)}\Big).
$$
The \gls{kld} error notion takes the one-hot-encoded target label as the first argument simplifying to
$
\error_{\mathcal{D}}^{KLD}(f) = 
    \mathbb{E}_{x,y\sim \mathcal{D}}
    [
    - \log f_y(x)
    ].
$
Interestingly, this error notion
only considers the softmax output of the ground truth class. 
In this sense, $\error_{\mathcal{D}}^{KLD}$ can be seen as a `soft' version of $\error_{\mathcal{D}}^{Top1}$ that incorporates uncertainty. \cref{fig:divergence_line} visualizes error notions. In comparison to \gls{hd} and \gls{jsd}, \gls{kld} has a steeper penalization for high uncertainties of the target class.

\section{Experimental study}
\label{sec:experimental_study}



\label{sec:experimental_setup}
We conduct experiments on CIFAR-10 \cite{cifar10} and  CIFAR-100 \cite{cifar100}.\footnote{Code made available at \url{https://github.com/monasch/divdis}}  
For the shift datasets, we use CIFAR-10C and CIFAR-100C \cite{hendrycks2019benchmarking} containing 18 types of synthetic corruptions in 5 severity levels. 
We employ 19 deep vision models pre-trained by \cite{huy_phan_2021_4431043} 
spanning a variety of vision models such as ResNet \cite{he2016deep} and VGG \cite{simonyan2014very}. 
In the \gls{tvt} scenario, we compute disagreement between all model pairs resulting in 171 disagreement instances. 

The \gls{tvc} setting measures disagreement of each vision model against the foundation model CLIP \cite{radford2021learning} finetuned on \gls{id} data. Here, CLIP serves as an anchor with superior generalisation abilities \cite{radford2021learning} possibly exposed to samples from $\mathcal{D}_{OOD}$ already during training. 
Model details are listed in \cref{app:testbed}. 


\input{plots/ontheline}

\paragraph{Does divergence disagreement pick up stronger on-the-line phenomena?}

We first inspect if the model's uncertainties incorporated in divergence disagreement and test error notions result in a stronger correlation between (i) ID vs. OOD disagreement (described as \textit{agreement-on-the-line} by \cite{baek2022agreement}) and (ii) ID vs. OOD test error (dubbed \textit{accuracy-on-the-line} by \cite{miller2021accuracy}). 
We compute pairwise disagreement and test error for the ID test sets of CIFAR-10/100 and the corrupted OOD test sets of CIFAR-10C/100C and plot ID vs. OOD values for each model pair. 

\cref{fig:ontheline} shows results exemplary for the fog corruption of CIFAR-10C.
We observe stronger correlations for divergence notions compared to the standard top-1 approach for both on-the-line phenomena in the setting with and without foundation model. As noticed by \cite{baek2022agreement}, the disagreement and test error line fit coincide. 
A stronger correlation is of interest because it could facilitate a more accurate line fit for estimating \gls{ood} test error, which will be our next point of discussion.


\input{tables/performance_estimation_tvt}

\paragraph{Can divergence disagreement provide better OOD performance estimation?}
To assess the potential of divergence disagreement for unlabeled OOD performance estimation, we employ a simple disagreement-based linear regression procedure, ALineD \cite{baek2022agreement}, and compare results across disagreement notions. 
AlineD exploits the observation that the slope and bias from the ID vs. OOD agreement matches the slope and bias from ID vs. OOD accuracy (see \cref{fig:ontheline}). Fitting regression coefficients to the disagreement relation and extrapolating with the ID test error thus gives an estimation for OOD test error without the need for labeled OOD data. Since the core requirement of the method is a correlation between ID and OOD agreement in the first place, we focus on CIFAR corruptions that exhibit such correlations ($R^2 > 0.95$) for all disagreement notions.

\cref{tab:aline} displays mean absolut percentage error (MAPE) of OOD performance estimation in the \gls{tvt} setting for datasets that satisfy the $R^2$ threshold.
We remark three main trends. Firstly, 
Hellinger disagreement provides the best OOD test error estimates on high correlation datasets reporting the smallest \gls{mape} in $17$ out of $22$ shifts. Secondly, \gls{hd} appears to be slightly less robust on the more labels containing CIFAR-100C:
Interestingly, for the same corruption (brightness 3+4) \gls{hd} disagreement performs best on CIFAR-10C but top-1 performs better on CIFAR-100C. This is indeed surprising, as one might expect divergence disagreement to prove particularly advantageous on datasets with many labels.
Thirdly, top-1 seems more robust on shifts with weak ID vs OOD disagreement: Considering all corruptions regardless of $R^2$, the median \gls{mape} for top-1  ($8.05$) is lower than for \gls{hd} ($10.15$), \gls{jsd} ($16.61$) and \gls{kld} ($19.62$) on CIFAR-10C. We present results for the \gls{tvc} scenario in \cref{sec:performance_estimation_fm}.


\input{plots/calibration}

\paragraph{How does miscalibration affect ID vs. OOD correlations?}
Next, we compare the impact of miscalibration on OOD performance estimation across disagreement notions. 
For that, we ask if the disagreement and test error correlation on ID vs. OOD data is more prone to miscalibration for divergence disagreement. This hypothesis sounds plausible since such notions rely heavily on the predictive distribution and could potentially explain poorer median estimates than top-1 (third observation from above). 
To assess this, we plot the class-aggregated calibration error (CACE) \cite{jiang2021assessing} averaged over all models in the testbed against the $R^2$ of the on-the-line phenomena for all corrupted datasets. We fit a three-degree polynomial per notion to highlight trends.  

\cref{fig:calibration} shows results. We highlight two main observations. First, the on-the-line phenomena vanish under increasing calibration error for all disagreement notions. This is an important insight since the question of when the phenomena are valid has been largely unexplained so far. Second, perhaps surprisingly, top-1 correlation is not more robust than divergence-based correlations. On CIFAR-100, we observe however, that Hellinger disagreement correlation drops faster with increasing calibration error (potentially explaining the second observation from above).

\paragraph{Can divergence disagreement detect OOD samples better?}
By comparing disagreement notions through the lens of OOD detection, we like to provide a different point of comparison. Unlike in OOD performance estimation, disagreement in OOD detection does not require a solid link to test error. Instead, it assumes distribution shifts affect models in random ways which lead them to disagree. We hypothesis that these perturbations affect all output dimensions equally and can thus be better captured by divergence disagreement. We use CIFAR-10 and CIFAR-100 test sets as ID samples and each "Hendrycks-corruptions" set as the OOD samples in separate OOD detection tasks. \Gls{msp} \cite{hendrycks2016baseline} and \gls{mlgt} \cite{hendrycks2019scaling} serve as baseline \gls{ood} scores.

\cref{tab:auc}  seems to confirm our hypothesis. It reports ROC-AUC for separating OOD from ID samples. Divergence disagreement performs best capturing information about the shift better than top-1, \gls{msp} and \gls{mlgt}. Interestingly, \gls{kld} performs best on this task, but worst on OOD error estimation indicating it picks up information about the distribution shift that could not be exploited for error estimation. 

\input{tables/ood_detection}



\label{sec:ood_detection}

\section{Conclusion}
By moving beyond the traditional top-class focus, we investigated more fine-grained notions of model disagreement that consider the full predictive label distribution. We show that contrasting models in a more nuanced way unlocks great potential for detecting system failure under distribution shift even more accurately. Future work may investigate the link to calibration more closely and sketch how model over- and under confidence affects disagreement. 

\section*{Acknowledgments}
We thank Metod Jazbec, Alexander Timans and Putra Manggala for helpful feedback on the draft. We are also grateful for the constructive comments from the anonymous reviewers that helped to improve this work. MS and EN are generously supported by the Bosch Center for Artificial Intelligence.

{\small
\bibliographystyle{plain}
\bibliography{egbib}
}

\clearpage
\appendix
\section{Testbed}
\label{app:testbed}
For the standard vision models, we use models pre-trained on CIFAR-10/CIFAR-100 by \cite{huy_phan_2021_4431043}. As foundation model, we finetune CLIP \cite{radford2021learning} on CIFAR \gls{id} data, which yield an accuracy increase on \gls{id} data from $88.8\%$ to $94.8\%$ on CIFAR-10 and from $61.7\%$ to $79.3\%$ on CIFAR-100 compared to the zeroshot version. We use a simple finetune procedure by adding a linear classification head to the image features outputted by CLIP. The classification head is trained for 10 epochs with a learning rate of $1\times 10^{-3}$.  \cref{tab:cifar10,tab:cifar100} summarize models, their test accuracy on ID data and parameter size. 

\input{tables/models_cifar10}
\input{tables/models_cifar100}

\newpage
\section{OOD performance estimation with CLIP}
\label{sec:performance_estimation_fm}

We evaluate performance estimation for both the \gls{tvt} and \gls{tvc} setting. In the \gls{tvc} scenario, we are interested in estimating performance of the standard vision models based on disagreement of each standard vision model with the finetuned CLIP model. We found the simpler line fitting method, ALine-S \cite{baek2022agreement}, to work better than ALine-D \cite{baek2022agreement} in this setting. ALine-D takes the disagreement of the model pair of interest directly into account (instead of indirectly through the fitted line coefficients, see \cite{baek2022agreement} for details). However, this yields to poor estimates when disagreement and error are far apart as observed in \cref{fig:ontheline}. 

\cref{tab:aline_fdm} reports results for datasets where agreement-on-the-line holds ($R^2 > 0.95$).  Interestingly, in this setting, \gls{jsd} and top-1 disagreement perform equally well and best on 7 shift datasets.

\input{tables/performance_estimation_tvc}

\end{document}

%% file: plots/divergence.tex
\begin{wrapfigure}{r}{0.55\linewidth}
  \centering
  \vspace{-8pt}
  \includegraphics[width=0.55\textwidth]{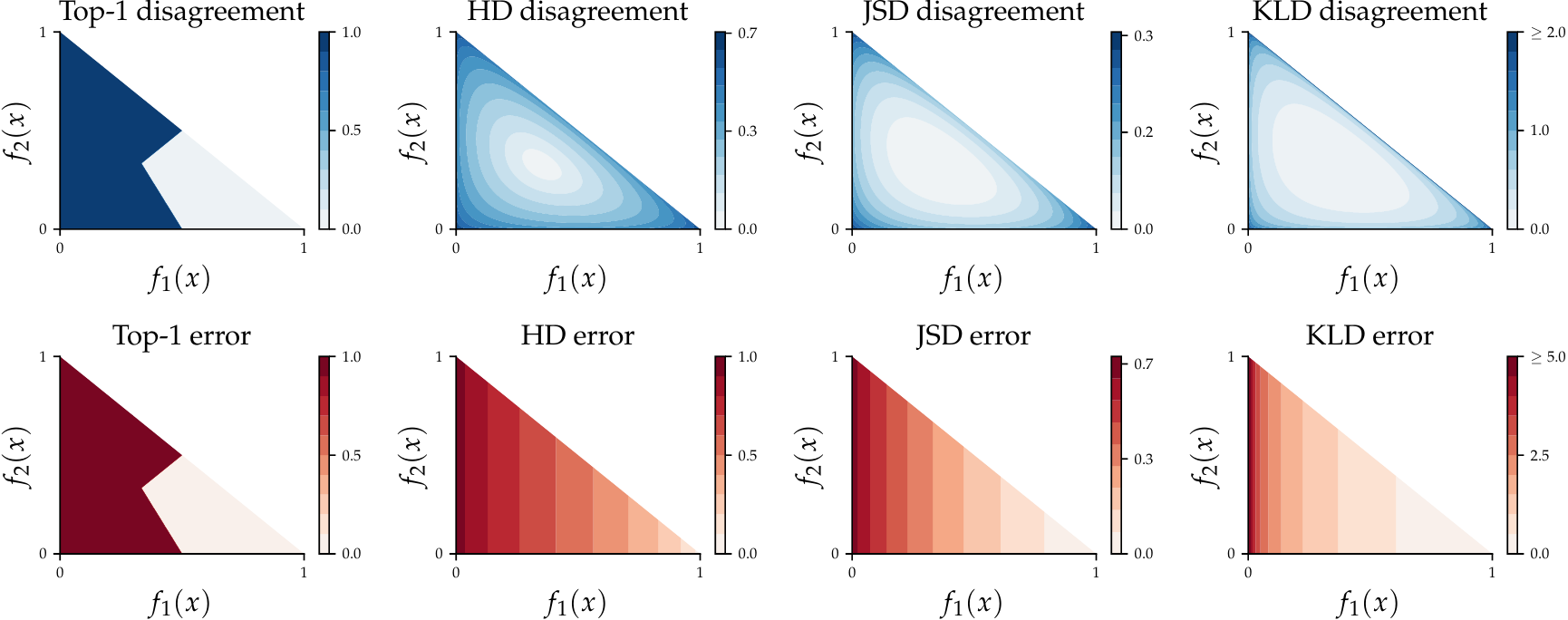}
  \caption{Disagreement and error notions for 3-class classification: $x$,$y$-axis represent predicted probability of model $f$ for class $1,2$. First row: disagreement between $f,f'$ with fixed $f'(x)=(0.35,0.325,0.325)$. Second row: error for $y=1$. 
  }
  \label{fig:divergence}
  \vspace{-10pt}
\end{wrapfigure}

%% file: plots/divergence_error.tex
\begin{wrapfigure}{r}{0.25\linewidth}
  \centering
  \vspace{-12pt}
  \includegraphics[width=0.25\textwidth]{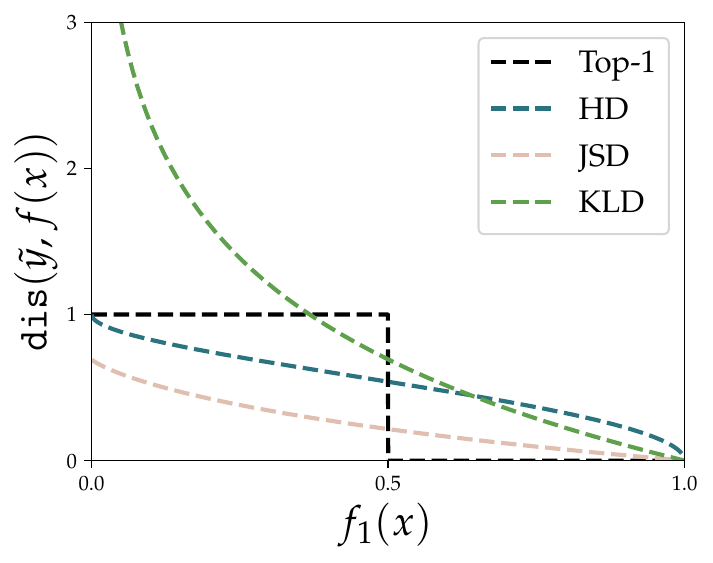}
  \caption{Error notions for binary classification for $y=1$\\ \\}
  \label{fig:divergence_line}
  \vspace{-35pt}
\end{wrapfigure}

%% file: plots/ontheline.tex
\begin{figure}
  \vspace{-15pt}
  \centering
  \includegraphics[width=\textwidth]{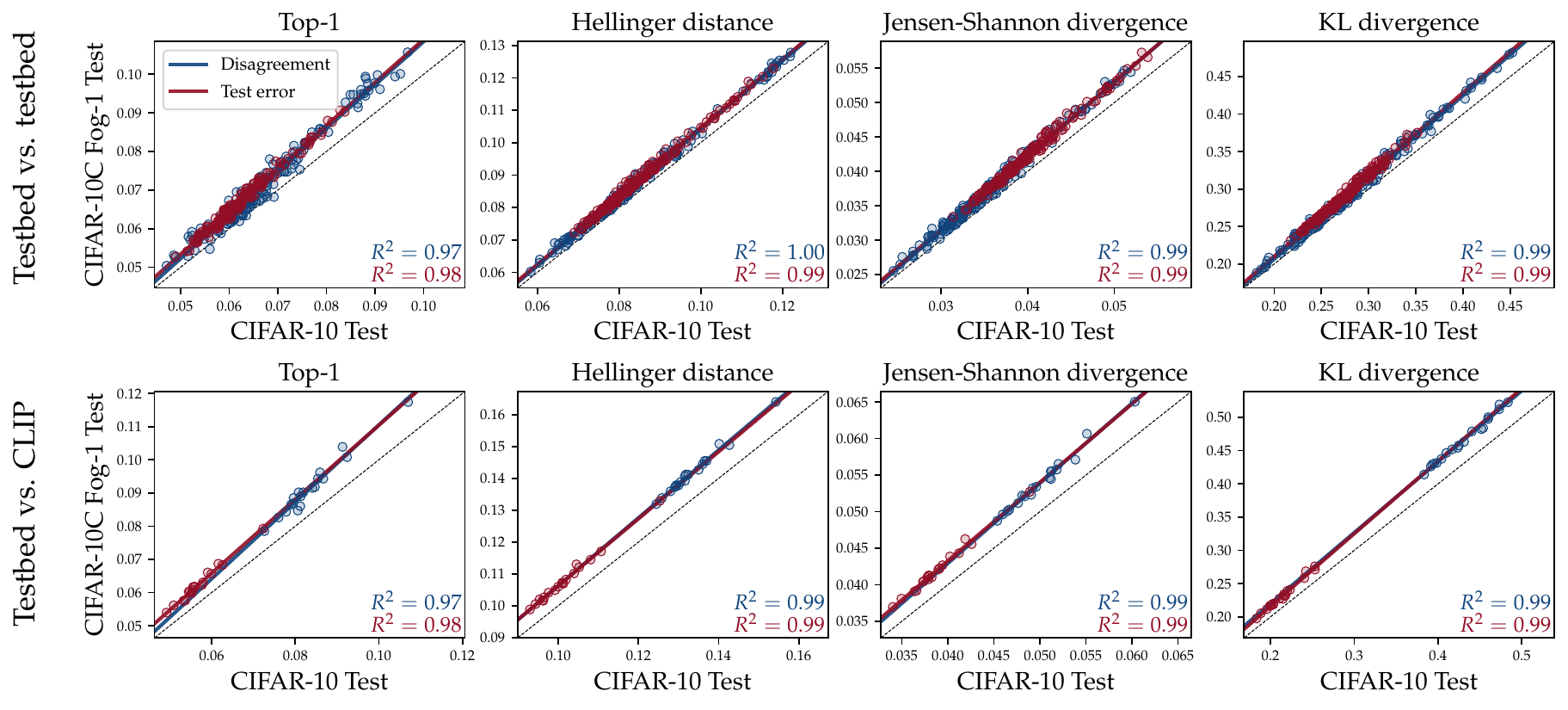}
  \caption{\textcolor{agreement}{\textit{Agreement-on-the-line}} and \textcolor{accuracy}{\textit{Accuracy-on-the-line}} for CIFAR-10C fog: Distributional disagreement notions (columns 2-4) correlate stronger than top-1 disagreement (first column). 
  }
  \label{fig:ontheline}
  \vspace{-0pt}
\end{figure}

%% file: tables/performance_estimation_tvt.tex
\begin{wraptable}{r}{7cm}
\vspace{-10pt}
\centering
\scalebox{0.80}{
\begin{tabular}{ll@{\hskip 15pt}c@{\hskip 15pt}c@{\hskip 15pt}c@{\hskip 15pt}c}
\toprule
& OOD dataset & Top-1 & HD & JSD & KLD \\
\midrule
\multirow{8}{*}{\rotatebox[origin=c]{90}{CIFAR-10C}} 
& brightness1    &              1.14 &            \textbf{0.39} &            0.76 &             0.83 \\
& brightness2    &              1.30 &            \textbf{0.47} &            0.99 &             1.03 \\
& brightness3    &              1.80 &            \textbf{0.81} &            1.23 &             1.57 \\
& brightness4    &              2.11 &            \textbf{1.26} &            1.70 &             1.71 \\
& contrast1      &              2.69 &            \textbf{1.49} &            1.70 &             3.77 \\
& defocus\_blur1  &              1.63 &            \textbf{0.58} &           0.91 &             1.31 \\
& fog1           &              2.26 &            \textbf{0.70} &            0.82 &             1.40 \\
& gaussian\_blur1 &              1.50 &            \textbf{0.58} &           0.90 &             1.31 \\
& saturate1      &              3.01 &            \textbf{2.90} &            4.80 &             6.77 \\
& saturate3      &              2.61 &            \textbf{1.22} &            2.12 &             2.91 \\
\midrule
\multirow{8}{*}{\rotatebox[origin=c]{90}{CIFAR-100C}} 
& brightness 1    &              0.63 &            \textbf{0.17} &            0.23 &             0.39  \\
& brightness 2    &              0.73 &            \textbf{0.41} &            0.43 &             0.71  \\
& brightness 3    &              \textbf{0.80} &            1.06 &            0.93 &             1.37  \\
& brightness 4    &              \textbf{0.94} &            2.10 &            1.93 &             2.82  \\
& contrast 1      &              1.28 &            \textbf{1.25} &            1.61 &             2.51  \\
& defocus blur 1  &              0.74 &            \textbf{0.46} &            0.63 &             0.91  \\
& defocus blur 2  &              \textbf{1.47} &            2.36 &            2.59 &             4.42  \\
& fog 1           &              0.86 &            \textbf{0.50} &            0.53 &             0.88  \\
& fog 2           &              \textbf{1.29} &            2.01 &            1.53 &             3.65  \\
& gaussian blur 1 &              0.80 &            \textbf{0.45} &            0.63 &             0.91  \\
& saturate 3      &              1.49 &            \textbf{1.28} &            1.34 &             3.03  \\
& saturate 4      &              \textbf{2.37} &            4.70 &            4.85 &             10.36  \\
\bottomrule
\end{tabular}
}
\caption{MAPE $(\downarrow)$ for OOD performance estimation: HD performs best overall, but is less performative under more labels. }
\label{tab:aline}
\vspace{-15pt}
\end{wraptable}

%% file: plots/calibration.tex
\begin{wrapfigure}{r}{0.5\linewidth}
        \vspace{-15pt}

        \centering        
        \includegraphics[width=0.5\textwidth]{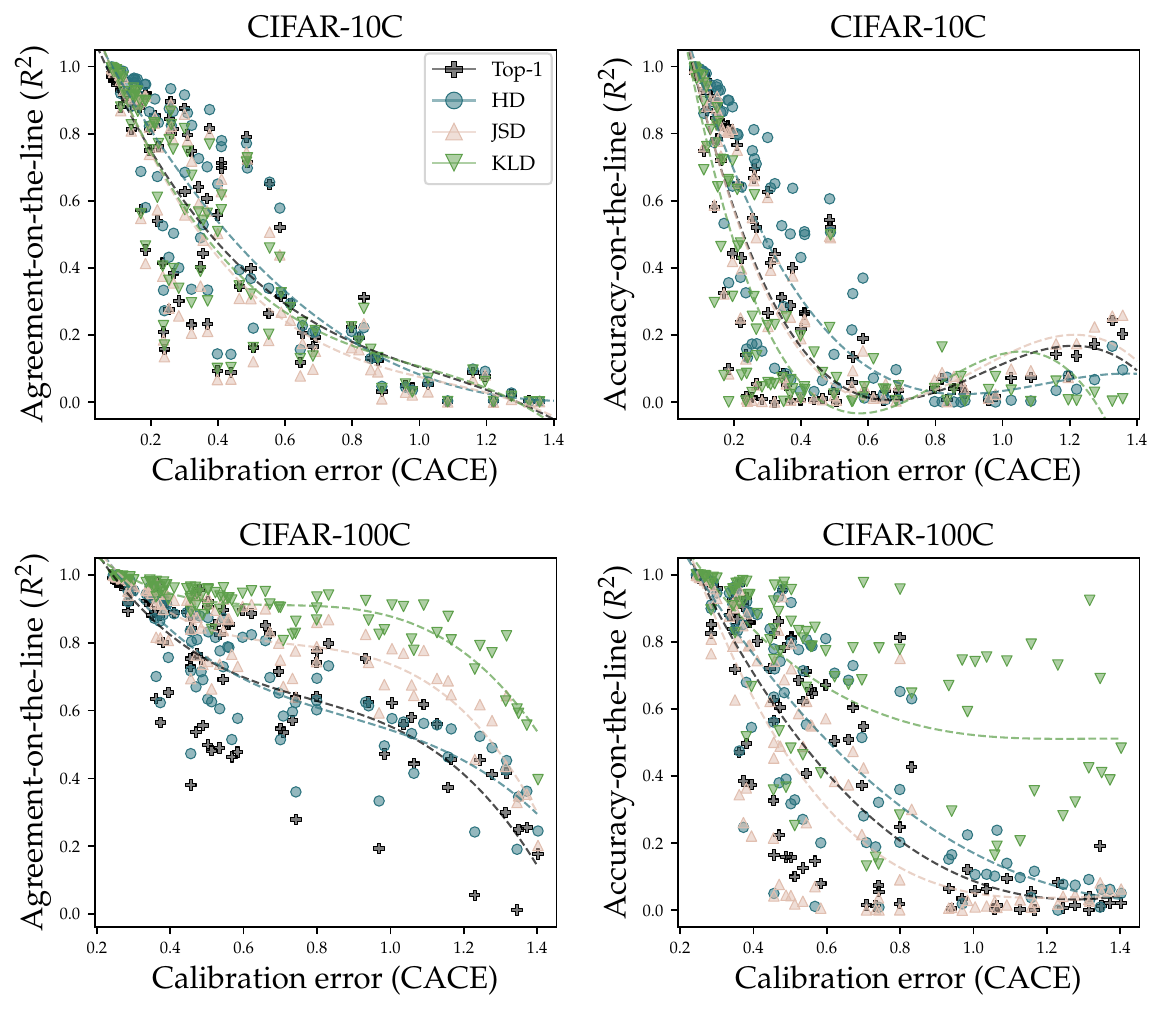}
        \caption{Calibration vs. agreement- and accuracy-on-the-line correlation ($R^2$): 
        Miscalibration reduces correlation.}
        \label{fig:calibration}
        \vspace{-20pt}

\end{wrapfigure}

%% file: tables/ood_detection.tex
\begin{table}[h]
\centering
\scalebox{0.9}{
\begin{tabular}{ll|rrrrrr}
\toprule
\multicolumn{2}{c}{Severity} & $-$MaxLogit& $-$MSP & Top-1 & HD & JSD & KLD \\
\midrule
\multirow{5}{*}{\rotatebox[origin=c]{90}{CIFAR-10C}} 
& 1        &            60.10 &           60.28 &         58.15 &       \textbf{60.47} &        60.33 &       60.46 \\
& 2        &            66.85 &           67.32 &         64.01 &       67.66 &        67.57 &       \textbf{67.67} \\
& 3        &            71.48 &           72.08 &         68.25 &       \textbf{72.46} &        72.34 &       72.44 \\
& 4        &            76.46 &           77.06 &         73.31 &       \textbf{77.59} &        77.42 &       77.54 \\
& 5        &            82.60 &           82.89 &         79.47 &       \textbf{83.52} &        83.26 &       83.34 \\
\midrule
\multirow{5}{*}{\rotatebox[origin=c]{90}{CIFAR-100C}} 
& 1        &            61.09 &           60.74 &         60.24 &       61.49 &        61.45 &       \textbf{61.80} \\
& 2        &            67.59 &           66.97 &         66.46 &       68.22 &        68.31 &       \textbf{68.59} \\
& 3        &            70.44 &           69.88 &         69.36 &       71.30 &        71.40 &      \textbf{71.66} \\
& 4        &            73.87 &           73.21 &         72.75 &       74.92 &        75.06 &       \textbf{75.37} \\
& 5        &            78.15 &           77.50 &         77.22 &       79.59 &        79.73 &       \textbf{80.10} \\
\bottomrule

\end{tabular}
}
\vspace{5pt}

\caption{ROC-AUC score for OOD detection in the \gls{tvt} scenario: Scores per severity level are averaged over all corruption types. HD and KLD disagreement detect OOD samples best. }

\label{tab:auc}
\end{table}

%% file: tables/models_cifar10.tex
\begin{table}[h]
    \centering
    \scalebox{0.8}{
    \begin{tabular}{l|cc}
    \toprule
     Model & Top-1 accuracy (\%) & Parameters (M) \\
    \midrule

    ResNet20 \cite{he2016deep} & 92.60 & 0.27 \\
    ResNet32 \cite{he2016deep} & 93.53 & 0.47 \\
    ResNet44 \cite{he2016deep} & 94.01 & 0.66 \\
    ResNet56 \cite{he2016deep} & 94.37 & 0.86\\
    VGG11 \cite{simonyan2014very} & 92.79 & 9.76 \\
    VGG13 \cite{simonyan2014very} & 94.00 & 9.94 \\
    VGG16 \cite{simonyan2014very} & 94.16 & 15.25 \\
    VGG19 \cite{simonyan2014very} & 93.91 & 20.57 \\
    MobileNetV2 (x0\_5) \cite{sandler2018mobilenetv2} & 92.88 & 0.70 \\
    MobileNetV2 (x0\_75) \cite{sandler2018mobilenetv2} & 93.72 & 1.37 \\
    MobileNetV2 (x1\_0) \cite{sandler2018mobilenetv2} & 93.79 & 2.24 \\
    MobileNetV2 (x1\_4) \cite{sandler2018mobilenetv2} & 94.22 & 4.33  \\
    ShuffleNetV2 (x0\_5) \cite{ma2018shufflenet} & 90.13 & 0.35 \\
    ShuffleNetV2 (x1\_0) \cite{ma2018shufflenet} & 92.98 & 1.26 \\
    ShuffleNetV2 (x1\_5) \cite{ma2018shufflenet} & 93.55 & 2.49 \\
    ShuffleNetV2 (x2\_0) \cite{ma2018shufflenet} & 93.81 & 5.37  \\
    RepVGG (a0) \cite{ding2021repvgg} & 94.39 & 7.84  \\
    RepVGG (a1) \cite{ding2021repvgg} & 94.89 & 12.82  \\
    RepVGG (a2) \cite{ding2021repvgg} & 94.98 & 26.82 \\\midrule
    CLIP ViT-B/32 \cite{radford2021learning} (finetuned) & 94.82 & 151.28 \\
    \bottomrule
    \end{tabular}
    }
    \vspace{5pt}
    \caption{Testbed for CIFAR-10}
    \label{tab:cifar10}
\end{table}

%% file: tables/models_cifar100.tex
\begin{table}[h]
    \centering
    \scalebox{0.8}{
    \begin{tabular}{l|cc}
    \toprule
    Model & Top-1 accuracy (\%) & Parameters (M) \\
    \midrule
    ResNet20 \cite{he2016deep} & 68.83 & 0.28  \\
    ResNet32 \cite{he2016deep} & 70.16 & 0.47  \\
    ResNet44 \cite{he2016deep} & 71.63 & 0.67  \\
    ResNet56 \cite{he2016deep} & 72.63 & 0.86  \\
    VGG11 \cite{simonyan2014very} & 70.78 & 9.80  \\
    VGG13 \cite{simonyan2014very} & 74.63 & 9.99  \\
    VGG16 \cite{simonyan2014very} & 74.00 & 15.30  \\
    VGG19 \cite{simonyan2014very} & 73.87 & 20.61  \\
    MobileNetV2 (x0\_5) \cite{sandler2018mobilenetv2} & 70.88 & 0.82  \\
    MobileNetV2 (x0\_75) \cite{sandler2018mobilenetv2} & 73.61 & 1.48  \\
    MobileNetV2 (x1\_0) \cite{sandler2018mobilenetv2} & 74.20 & 2.35  \\
    MobileNetV2 (x1\_4) \cite{sandler2018mobilenetv2} & 75.98 & 4.50  \\
    ShuffleNetV2 (x0\_5) \cite{ma2018shufflenet} & 67.82 & 0.44  \\
    ShuffleNetV2 (x1\_0) \cite{ma2018shufflenet} & 72.39 & 1.36  \\
    ShuffleNetV2 (x1\_5) \cite{ma2018shufflenet} & 73.91 & 2.58  \\
    ShuffleNetV2 (x2\_0) \cite{ma2018shufflenet} & 75.35 & 5.55  \\
    RepVGG (a0) \cite{ding2021repvgg} & 75.22 & 7.96 \\
    RepVGG (a1) \cite{ding2021repvgg} & 76.12 & 12.94 \\
    RepVGG (a2) \cite{ding2021repvgg} & 77.18 & 26.94 \\
    \midrule
    CLIP ViT-B/32 \cite{radford2021learning} (finetuned) & 79.28 & 151.28 \\
    \bottomrule
    \end{tabular}
    }
    \vspace{5pt}
    \caption{Testbed for CIFAR-100}
    \label{tab:cifar100}
\end{table}

%% file: tables/performance_estimation_tvc.tex
\begin{table}[h]
\centering
\scalebox{0.8}{
\begin{tabular}{ll@{\hskip 15pt}c@{\hskip 15pt}c@{\hskip 15pt}c@{\hskip 15pt}c}
\toprule
& OOD dataset & Top-1 & HD & JSD & KLD \\
\midrule
\multirow{8}{*}{\rotatebox[origin=c]{90}{CIFAR-10C}} 
& brightness 1    &              1.30 &            \textbf{0.47} &                0.62 &              0.53 \\
& brightness 3    &              \textbf{2.48} &            3.86 &                3.92 &              3.87 \\
& brightness 4    &              \textbf{4.62} &            9.17 &                6.96 &              7.92 \\
& contrast 1      &              5.07 &            5.09 &                \textbf{3.17} &              4.03 \\
& defocus blur 1  &              \textbf{1.95} &            6.44 &                3.67 &              4.65 \\
& fog1           &              1.97 &            1.96 &                 \textbf{1.76} &              2.98 \\
& gaussian blur 1 &              \textbf{2.16} &            6.68 &                3.76 &              4.83 \\
& saturate 3      &              2.60 &            1.64 &                \textbf{1.62} &              2.19 \\
\midrule
\multirow{8}{*}{\rotatebox[origin=c]{90}{CIFAR-100C}} 
& brightness 1    &              0.63 &            \textbf{0.22} &                0.30 &              0.32 \\
& brightness 2    &              0.99 &            0.59 &                \textbf{0.56} &              1.01 \\
& brightness 3    &              \textbf{1.20} &            1.54 &                0.96 &             3.05 \\
& contrast 1      &              1.36 &            1.14 &                \textbf{0.97} &              2.62 \\
& defocus blur 1  &              1.10 &            0.59 &                \textbf{0.40} &              0.70 \\
& fog1           &              1.71 &            0.73 &                0.83 &             \textbf{0.67} \\
& gaussian blur 1 &              1.24 &            0.55 &                \textbf{0.40} &             0.74 \\
& saturate 3      &              \textbf{1.08} &            2.17 &                1.31 &             4.59 \\
& saturate 4      &              \textbf{1.29} &            7.98 &               5.38 &             15.44 \\
\bottomrule
\end{tabular}
}
\vspace{5pt}

\caption{MAPE $(\downarrow)$ for OOD performance estimation of vision models using disagreement against a finetuned CLIP model (\gls{tvc} setting).}

\label{tab:aline_fdm}
\vspace{-20pt}
\end{table}